\pdfoutput=1
\documentclass[11pt]{article}

\usepackage[final]{acl} % MODIFICATION: Ensure this is the correct style (review/final/preprint) and package for your venue.

\usepackage{times}
\usepackage{latexsym}

\usepackage[T1]{fontenc}
\usepackage[utf8]{inputenc}

\usepackage{microtype}

\usepackage{inconsolata}

\title{Social Hatred: Efficient Multimodal Detection of Hatemongers} % MODIFICATION: Slightly shortened title suggestion if needed. Original: Instructions for *ACL Proceedings
\author{Tom Marzea \\
    Ben Gurion University\\
  \texttt{tommarz@post.bgu.ac.il} \\\And
  Abraham Israeli \\
  University of Michigan\\
  \texttt{isabrah@umich.edu} \\\And
  Oren Tsur \\
  Ben Gurion University\\
  \texttt{orentsur@bgu.ac.il}
  }

\usepackage{times}  % DO NOT CHANGE THIS
\usepackage{helvet}  % DO NOT CHANGE THIS
\usepackage{courier}  % DO NOT CHANGE THIS
\usepackage{graphicx} % DO NOT CHANGE THIS
\usepackage{natbib}  % DO NOT CHANGE THIS AND DO NOT ADD ANY OPTIONS TO IT
\usepackage{caption} % DO NOT CHANGE THIS AND DO NOT ADD ANY OPTIONS TO IT
\usepackage{algorithm}
\usepackage{algorithmic}

\usepackage{newfloat}
\usepackage{listings}
\DeclareCaptionStyle{ruled}{labelfont=normalfont,labelsep=colon,strut=off} % DO NOT CHANGE THIS
\floatstyle{ruled}
\newfloat{listing}{tb}{lst}{}
\floatname{listing}{Listing}
\usepackage{tabularx}
\usepackage{makecell}
\usepackage{adjustbox}
\usepackage{bm}

\usepackage{mathtools}
\usepackage{multirow}

\usepackage{pifont}
\newcommand{\cmark}{\ding{51}}%
\newcommand{\xmark}{\ding{55}}%
\newcommand{\crossedcmark}{\textcolor{black}{\ding{51}}{\small\textcolor{black}{\kern-0.7em\ding{55}}}}
\usepackage{tikz,graphics,color,epsf,subcaption} % MODIFICATION: Ensure all these are needed and allowed. `graphics` is usually superseded by `graphicx`. `color` might be disallowed in text. `epsf` is quite old.
\usepackage{xcolor} % MODIFICATION: Ensure color package is allowed if used in main text (checklist might be appendix).

\begin{document}

\maketitle

\begin{abstract}
Automatic detection of online hate speech serves as a crucial step in the detoxification of the online discourse. Moreover, accurate classification can promote a better understanding of the proliferation of hate as a social phenomenon.
While most prior work focus on the detection of hateful \emph{utterances}, we argue that focusing on the \emph{user} level is as important, albeit challenging. In this paper we consider a multimodal aggregative approach for the detection of hate-mongers, taking into account the potentially hateful texts, user activity, and the user network.
Evaluating our method on three unique datasets X (Twitter), Gab, and Parler we show that processing a user's texts in her social context significantly improves the detection of hate mongers, compared to previously used text and graph-based methods. We offer comprehensive set of results obtained in different experimental settings as well as qualitative analysis of illustrative cases.
Our method can be used to improve the classification of coded messages, dog-whistling, and racial gas-lighting, as well as to inform intervention measures. Moreover, we demonstrate that our multimodal approach performs well across very different content platforms and over large datasets and networks.
\end{abstract}

{\bf Offensive content warning:} The illustrative examples throughout the manuscript, and specifically in Table
\ref{tab:hate_texts} may be offensive to some readers. 

\section{Introduction}
\label{sec:intro}
The rising popularity of social platforms coincides with the proliferation of online hate speech and a surge in hateful content targeting minorities \cite{waseem2016hateful,laub2019hateGlobalComparison}. Accordingly, there is a growing body of research on the appearance and magnitude of hate speech on social media \cite{knuttila2011user, chandrasekharan2017you, zannettou2018gab}, and on hate speech detection \cite{saleem2017web, waseem2016hateful, davidson2017automated}. Online hate is often linked to real-world violence \cite{altrightpipeline,violence2019,8chan,mcilroy2019welcome, mathew2019spread,allenShooting23}. 

Hate is often promoted by communities, rather than by isolated individuals. Shifting the focus from the utterance (post) level to the user and community level can be beneficial: it may provide a better understanding of hate group dynamics; it could improve detection of coded language (dog whistling, gas-lighting) at the post level; and it could inform intervention strategies\footnote{Intervention alternatives depend on legal regulation and platform policy and range from the removal of a specific post, the flagging of a specific post, to the temporal or indefinite suspension of an account.} \cite{thomas2023disrupting}. 

Challenges in utterance-level hate speech classification are well-documented \cite{Talat2018,fortuna-etal-2022-directions,velankar2022review, kim2024labelaware}, and even state-of-the-art LLMs show shortcomings in nuanced social tasks like hate detection \cite{ziems24llms4css}. 

Table \ref{tab:hate_texts} provides examples illustrating these challenges. While explicit hate (utterance \#1) is often detected, coded language and symbols (\#2, the (((echo))) symbol\footnote{\url{www.adl.org/resources/hate-symbol/echo}}) or white supremacist slogans (\#3, the `14 words'\footnote{\url{www.adl.org/resources/hate-symbol/14-words}}) may require external knowledge or specific training data. Allusions (\#4-7) and twisted slogans (\#7-8) further complicate detection for both models and humans. A detailed analysis of these examples is provided in Appendix \ref{sec:table1_analysis}. 

\begin{table*}[ht!]
\footnotesize
\centering
\begin{tabular}{c|m{8.5cm}|c|c|c|c|c }
 & \multicolumn{1}{c|}{Text} & DHB & JS & FTDB & GPT4 & Gemini 1.5Pro \\ \hline\hline
1  & ``Muslim terrorists did this and obama brought them here to run our government. Gotta get rid of Obammy, Omar and all these other f**king Muslims'' & .57 & .9 & .98 & \cmark & \cmark\\ \hline
2  & ``is there a more debased creature in our modern pop-cultural cesspool than (((Amy Schumer)))? Probably not.'' & .22 & .35 & .92  &\cmark $^?$  & \cmark\\\hline
3 &   ``We must secure the existence of our people and a future for white children!'' & .03 & .36 & .02 & \cmark & \cmark $^?$ \\ \hline
4  &  ``If white people are going to continue  existing - we must support our race''  & .06 & .4 & .69 &\cmark $^?$  & \cmark $^?$  \\ \hline
5  &  ``When the youngest white generation becomes adult they will be 5\% of the world population, down from 1/3 in 1900'' & .06 & .25 & .01  &\xmark & \cmark $^?$ \\ \hline
6  & ``Did you know that most dog breeds were bred by whites? We’re awesome and they’re awesome!'' & .2 & .25 & .65  &\xmark & \xmark \\ \hline
7 &  ''White lives matter!'' & .02 & .24 & .02  &\xmark  & \xmark $^?$ \\ \hline
8 &  ``Blue lives matter!'' & .02 & .01 & .01  &\xmark & \xmark $^?$ \\\hline
\end{tabular}
\caption{Examples of hate-promoting texts. DHB: deHateBERT \cite{aluru2020deep}; JS: Google's Jigsaw; FTDB: DistillBERT fine-tuned on our datasets; GPT-4 \& Gemini 1.5 Pro predictions (see Appendix \ref{sec:generated_predictions}). `?' indicates nuanced prediction.} % MODIFICATION: Slightly shortened caption.
\label{tab:hate_texts}
\end{table*}

Contextual information and aggregation at the user level can help mitigating challenges posed by individual posts. Multiple implicit posts from one user, or association with other hateful users/posts, can reinforce weak signals. However, aggregation requires careful consideration of prediction confidence and context, e.g., how many implicit posts equate to one explicit post? We propose a principled way to combine predictions and modalities for accurate user-level classification.

We explore three aggregation approaches: (i) binary weights with a fixed threshold, (ii) relational aggregation using social context, and (iii) distributional aggregation based on confidence levels. We combine these into a multimodal model.

\paragraph{Contribution} The contribution of this work is threefold:
\vspace{-.2cm} 
\begin{enumerate}
    \item We propose a robust and efficient multimodal aggregative approach for detecting hate-mongers.
    \vspace{-.2cm}
    \item We demonstrate the benefits of contextual aggregation over three unique datasets (Twitter, Gab, Parler).
    \vspace{-.2cm}
    \item We share a novel annotated dataset of Parler hate users. 
\end{enumerate}
\vspace{-.2cm}

\section{Related Work}
\label{sec:related_work}
Hate speech detection methods and challenges are surveyed by \citet{alkomah2022}, \citet{elsherief2021latent,velankar2022review,fortuna-etal-2022-directions}. Issues include subjectivity \cite{khurana-etal-2022-hate}, limitations of transfer learning \cite{israeli2022free}, and the need for robust evaluation \cite{rottger2020hatecheck,rottger2022multilingual}. Implicit hate taxonomies exist but the datasets are often unsuitable for user-level analysis \cite{elsherief2021latent}. % MODIFICATION: Condensed paragraph on general HS detection.

Research is shifting towards the user level. Early work explored demographics \cite{waseem2016hateful} and account meta-features \cite{ribeiro2017like}. \citet{arviv2021sa} used a multi-modal approach combining predictions from a user's posts, followers, and followees. % MODIFICATION: Condensed paragraph on early user-level work.

Textual signals combined with network diffusion models have been used to propagate initial hate labels within communities of users \cite{ribeiro2018characterizing, israeli2022free}. Graph Neural Networks (GNNs) incorporate text and network structure \cite{li2021neighbours, miao22, das2021you, Utku2024}.  GNNs were used for  detection of hate on post-level \cite{miao22} as well as for classification on the user level \cite{das2021you, Utku2024}. Early detection of potential spreaders using author profiling techniques has also been explored \cite{Irani2021EarlyDO}. % MODIFICATION: Condensed paragraph on network/GNN approaches.

Recent work explores using LLM rationales for interpretability \cite{nirmal2024interpretable}, though fine-tuned models often outperform LLMs on socially nuanced tasks like hate detection \cite{ziems24llms4css}. % MODIFICATION: Condensed paragraph on LLMs.

%==============================================================================

\section{Multimodal Aggregative Approaches}
\label{sec:methods}

\subsection{Aggregative Approaches}
\label{subsec:aggregative}
Our multimodal framework makes relies on two modalities: the textual modality and the social modality. The user-level classification is achieved by aggregating predictions for multiple utterances made by a specific user, informed by the user social context. We therefore describe the classification of of a single utterance, before offering a number of ways to aggregate predictions in order to achieve accurate classification at the user level. 

\paragraph{\bf Utterance-level Classification ($C^T$)} The foundation is single-utterance classification. Any model outputting a probability $\theta(t)$ can be used. We binarize using a threshold $\tau^T$:
%\vspace{-.2cm} % MODIFICATION: Removed negative vspace
\begin{align}
    C^T(t) = %PLM(t,M,\tau^T)=
            \begin{cases}
            1& {\theta(t)}\geq \tau^T\\
            0& \text{otherwise}
        \end{cases}
        \label{eq:plm}
\end{align}
%\vspace{-.2cm}
We use a fine-tuned DistilBERT \cite{sanh2019distilbert} following \citet{israeli2022free}.

\paragraph{\bf User-level Classification ($C^U$)}
\label{subsec:user_level}
Detecting hateful users involves aggregating signals from their posts ($T^u$). The generic form of $C^U$ uses an aggregation function $\Theta(u)$ and a user threshold $\tau^U$:
%\vspace{-.2cm}
\begin{align}
    C^U(u) =
            \begin{cases}
            1& {\Theta(u)\geq \tau^U}\\
            0& \text{otherwise}
        \end{cases}
        \label{eq:ulm}
\end{align}
%\vspace{-.2cm}

\paragraph{\bf Naive Aggregation with Fixed Threshold ($\Theta_F$)} A simple approach counts hateful posts by user $u$:
\begin{align}
    \Theta_F(u) = \sum_{t\in T^u}{C^T(t)}
    \label{eq:fixed}
\end{align}
The threshold $\tau^U$ controls sensitivity (e.g., $\tau^U=1$ for zero-tolerance\footnote{Often too harsh: remember that in the standard case $\tau^T=0.5$ thus a user will be labeled a hate-monger even if $\theta(t)=0.51$ for one of her posts and $\theta(t')< 0.5 ~~~\forall t' \in T^u / \{t\}$.}, higher values for repeated offenders). We denote this classification $C_F$.

We propose more robust aggregations considering nuance and social context, training $\Theta$ (e.g., logistic regression) on user features derived from these aggregations.

\paragraph{Multimodal Relational Aggregation ($\Theta_R$)} User identity is shaped by social ties \cite{mcmillan1986sense,wellman1999virtual}, at time including affiliation with online hate groups \cite{gordon2017second,govers2023down}. We incorporate the hate level in a user $u$'s ego network (followers $\overleftarrow{u}$, followees $\overrightarrow{u}$) in graph $G(V,E)$. $\Theta_R(u)$ combines $u$'s own post count ($C_F(u)$) with the proportion of hateful neighbors:
%\vspace{-.2cm}
\begin{multline}
\Theta_R(u) = \alpha \cdot C_F(u) + \beta \cdot \frac{1}{|\overleftarrow{u}|}\sum_{v\in{\overleftarrow{u}}} C_F(v) + \\ \gamma \cdot \frac{1}{|\overrightarrow{u}|}\sum_{v\in{\overrightarrow{u}}} C_F(v)
\label{eq:relational} % MODIFICATION: Added label for potential reference
\end{multline}
%\vspace{-.2cm}
Weights $\alpha, \beta, \gamma$ are learned.

\paragraph{\bf Distributional Aggregation ($\Theta_D$)} This approach addresses variations in hate intensity (commitment, implicitness). Instead of just counting hateful posts, we consider the distribution of hate scores $\theta(t)$ for user $u$'s posts $T^u$. We represent this distribution as a $k$-dimensional vector using bins or quantiles.
For bin-based, the [0,1] range is split into $k$ bins; the vector counts posts falling into each bin. For quantile-based, the range $[\min(\theta(t)), \max(\theta(t))]$ for $t \in T^u$ is split into $k$ bins.
We optimize:
%\vspace{-.2cm}
\begin{align}
    \Theta_D(u) = \sum_{i=1}^k w_i \cdot \sigma(B_i(u))
    \label{eq:distributional}
\end{align}
where $B_i(u)$ is the count for bin $i$, $\sigma$ is softmax, and $w_i$ are learned weights.

\paragraph{Combined Multimodal Aggregation ($\Theta$)} We combine relational and distributional features. Using $\Theta^b_D$ (bin-based) and $\Theta^q_D$ (quantile-based):
%\vspace{-.2cm}
\begin{align}
    \Theta(u) = \Theta^b_D(u) + \Theta^q_D(u) + \Theta_R(u) % MODIFICATION: Corrected typo (missing (u))
    \label{eq:multimodal}
\end{align}
The final classifier learns weights for all components.

\subsection{Socially-aware Baselines}
\label{subsec:baselines}
We compare the multimodal aggregations against a number of strong baselines that leverage the network structure. 

\paragraph{\bf DeGroot's Diffusion} A belief propagation model shown effective for hate detection \cite{ribeiro2018characterizing,israeli2022free}.

\paragraph{\bf Graph Neural Networks (GNNs)} Methods explored by \citet{das2021you} and others for hateful user detection:
\begin{enumerate}
    \item {\bf Node2Vec} \cite{grover2016node2vec}: Embeddings from biased random walks. % 
    \item {\bf GCN} \cite{kipf2016semi}: Convolutional layers aggregating neighbor features. % 
    \item {\bf GAT} \cite{velickovic2017graph}: Attention mechanism weighting neighbor importance. 
    \item {\bf GraphSAGE} \cite{hamilton2018inductive}: Inductive learning via sampling and aggregating neighbor features. 
    \item {\bf{AGNN}} \cite{thekumparampil2018attentionbased}: Attention-based propagation learning adaptive local summaries. 
\end{enumerate}
AGNN and GCN were specifically used for hate user detection by \cite{das2021you} and \cite{Utku2024}.

%====================================================================================

\section{Datasets and Annotation}
\label{sec:data}

We use three datasets: Twitter-Echo, Gab, and Parler. Descriptive statistics are provided in Tables \ref{tab:table_datasets2} and \ref{tab:table_datasets4}.

\begin{table*}[ht!] 
    \centering
    {\small
        \begin{tabular}{c@{\quad}|c@{\quad}|c@{\quad}c@{\quad}|c@{\quad}c@{\quad}|c@{\quad}c@{\quad}}
          \multicolumn{1}{c}{} & \multicolumn{1}{c}{} & \multicolumn{2}{c}{Raw Data} & \multicolumn{4}{c}{Annotated Data (Overall)}\\
          {Dataset} & {Source} & {\#Posts} & {\#Users} & {\#Posts} & {\% Hate } & {\#Users} & {\% Hate}\\
          \hline\rule{0pt}{12pt}
            Echo (Twitter) & \citet{arviv2021sa} & 18M & 7.1K & 4.6K & 8.2\% & 1K & 15.4\%\\
            Gab & \citet{alma9926823946704361} & 22M & 337K & 5K & 5.1\% & 1K & 24.8\%\\
            Parler & This work (users) & $183M^\dagger$ & $4.1M^\dagger$ & 8.3K$^\ddagger$ & 32.9\%$^\ddagger$ & 890 & 25.4\% \\
          \hline
          %\multicolumn{8}{l}{\footnotesize $\dagger$ Raw data from \cite{aliapoulios2021early}. $\ddagger$ Post annotations from \cite{israeli2022free}. User level annotations are introduced in this work.}
        \end{tabular}
    }
    \caption{Overview of datasets and annotated subsets. $\dagger$ Raw data from \cite{aliapoulios2021early}. $\ddagger$ Post annotations from \cite{israeli2022free}. User level annotations are introduced in this work.}
    \label{tab:table_datasets2} 
\end{table*}

\begin{table*}[ht!]
    \centering
    {\small
        \begin{tabular}{c@{\quad}|c@{\quad}c@{\quad}c@{\quad}|c@{\quad}c@{\quad}|c@{\quad}c@{\quad}}
          \multicolumn{1}{c}{} & \multicolumn{5}{c|}{Largest Connected Component (LCC) Stats} & \multicolumn{2}{c}{Annotated in LCC}\\
          {Dataset} & {\#Posts} & {\#Users} & {\#Edges} & {Clust. Coeff.} & {Gamma ($\gamma$)} & {\#Users} & {\% Hate}\\
          \hline\rule{0pt}{12pt}
            Echo (Twitter) & 9.8M & 3.7K & 20.7K & 0.190 & 2.8 & 532 & 26.1\%\\
            Gab & 19.3M & 51.2K & 2.47M & 0.402 & 4.1 & 982 & 24.5\%\\
            Parler & 60.7M & 643K & 11.4M & 0.224 & 2.1 & 881 & 25.2\%\\
          \hline
          \multicolumn{8}{l}{\footnotesize Clustering Coefficient and Optimal Gamma (scale-free fit) calculated on undirected graph.}
        \end{tabular}
    }
    \caption{Statistics for the largest (weakly) connected component used in experiments.}
    \label{tab:table_datasets4} 
\end{table*}

\paragraph{\bf Echo (Twitter)}
\label{sec:data_echo}
The triple parentheses, or triple brackets, also known as the (((echo))), is an antisemitic symbol that is used to highlight the names of individuals of Jewish background (e.g., actress and comedian Amy Schumer, see utterance \#2 in Table \ref{tab:hate_texts}), organizations owned by Jewish people (e.g., Ben \& Jerry's), or organizations accused of promoting ``Jewish globalist values'' (e.g., the International Monetary Fund).
The Echo dataset curated by \citet{arviv2021sa} contains over 18M English tweets posted by $\sim$7K echo users between May and June 2016. Annotations are provided at the tweet and the user level. An important feature of this dataset is that all users have utterances containing the echo symbol, although some users use it in a non-hateful manner, e.g., to symbolize a hug. This ambiguous nature of the symbol makes hate detection challenging. 
%Focuses on the antisemitic (((echo))) symbol used ambiguously \cite{arviv2021sa}. Contains 18M tweets from ~7K users (May-June 2016). Tweet and user annotations provided. All users in the data used the echo symbol, making detection challenging. % MODIFICATION: Condensed description.

\paragraph{\bf Gab}
\label{sec:data_gab}
Positioning itself as putting ``people and free speech first'', Gab welcome users suspended from other social networks. The platform permits pornographic and obscene content, as long as it is labeled \emph{NSFW} (`not safe for work'). Posts (called \emph{gabs}) are limited to 300 characters, and users can repost, quote, or reply to previously created gabs. 
The dataset (22M posts, 337K users) was collected by \cite{zannettou2018gab} from Aug 2016 - Jan 2018. An annotated subset shared by \citet{alma9926823946704361} includes ~60K posts and 1K users. 

\paragraph{\bf Parler}
\label{sec:data_parler}
Alluding to the French verb `to speak', Parler was launched on August 2018. On April 2023, the platform was acquired by Starboard and was taken offline to ``undergo a strategic assessment'' (Starboard announcement on Parler's landing page \url{https://parler.com/}, accessed: 5/8/2023). The platform was relaunched in February 2024, announcing it is ``breaking free from the constraints of conventional platforms''. 

The platform branded itself as ``The World's Town Square''  a place to \emph{``Speak freely and express yourself openly, without fear of being ``deplatformed'' for your views''}.\footnote{Parler branding on its landing page.} 
Parler's official guidelines\footnote{\url{https://parler.com/documents/guidelines.pdf}} explicitly allowed ``trolling'' and ``not-safe-for-work'' (NSFW) content, including only three ``principles'' prohibiting ``unlawful acts'', citing ``Obvious examples include: child sexual abuse material, content posted by or on behalf of terrorist organizations, intellectual property theft''. Parler allows posts of up to 1000 characters, compared to 280/300 allowed by Twitter/Gab.

A dataset of Parler messages was introduced by \citet{aliapoulios2021early} and a subset of 10K \emph{posts} was annotated by \citet{israeli2022free}. As part of this research, we share the first annotated dataset of Parler \emph{users}.

\paragraph{\bf Annotation of Parler Users}  Annotation was done by 94 senior year Data Science B.Sc students for bonus course credit. Annotators were introduced to Parler and were given explicit instructions about the annotation task. The annotation process involved rating each account on a 1--5 scale (non-hateful -- extremely hateful). We ensured that each user was annotated by three annotators. The full annotation guidelines and further details regarding the sampling method and the annotation protocol are available in Appendix \ref{annotation_guidelines}.

%==============================================================================dix

\section{Results and Analysis}
\label{sec:user_level_results}

\subsection{Experimental Settings}
\label{subsec:user_level_experiment}

Utterance-level: We fine-tuned DistilBERT \cite{Sanh2019DistilBERTAD} per dataset (80/20 train/val split, batch 32, early stopping). 
User-level: 5-fold CV on the largest connected component (LCC) of each dataset (see details in  Table \ref{tab:table_datasets4}). GNN settings follow \citet{das2021you}. DeGroot settings follow \citet{israeli2022free}. Node2Vec used implementation by \citet{btab2021}. Further details about models' configurations, hyperparameters, and application can be found in Appendix \ref{ap:exp_details}. % MODIFICATION: Moved detailed hyperparams/settings to appendix.

\subsection{Results and Analysis}

\begin{figure}[ht!]
    \centering
    \includegraphics[width=.4\textwidth]{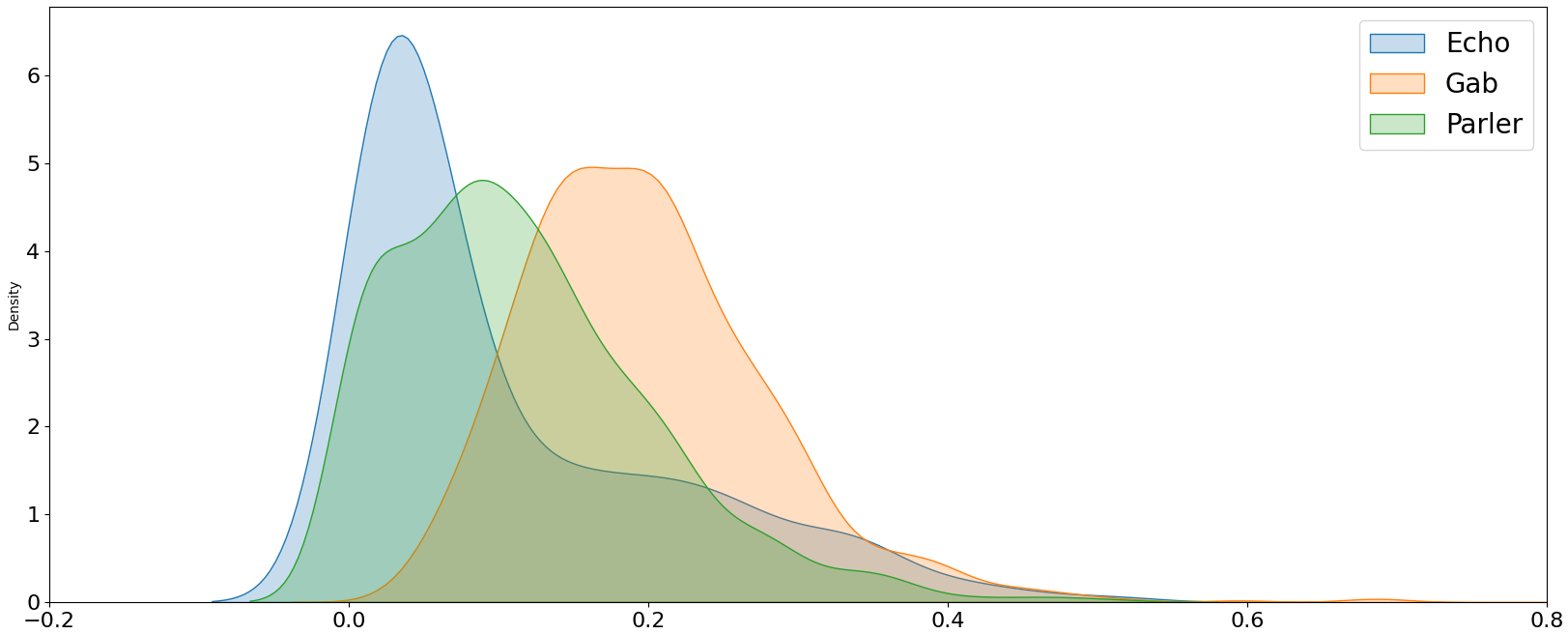} % MODIFICATION: Consider reducing width further if needed, e.g., 0.38\textwidth
    \caption{Density of mean utterance scores per user.} % MODIFICATION: Shortened caption
    \label{fig:post_mean_probas_per_user_kde}
\end{figure}

\paragraph{\bf Utterance Level Prediction} Vanilla DistilBERT performance (Table \ref{tab:post_level_module_metrics}) reveal the challenge in the detection of hate on the utterance level, especially on Gab (F1=0.29), motivating user-level aggregation. We observe that  mean hate score distributions differ across platforms (Figure \ref{fig:post_mean_probas_per_user_kde}), revalidating the need for a generic yet robust approach.

\begin{table}[ht!]
    \centering
    \small % MODIFICATION: Added \small
    \begin{tabular}{l|c|c|c}
        \hline
        \rule{0pt}{10pt}Dataset & Precision & Recall & F1 Score \\ % MODIFICATION: Reduced rule height slightly
        \hline\rule{0pt}{10pt}Echo & 0.412 & 0.803 & 0.545 \\ % MODIFICATION: Rounded
        Gab & 0.206 & 0.547 & 0.299 \\ % MODIFICATION: Rounded
        Parler & 0.632 & 0.818 & 0.713 \\ % MODIFICATION: Rounded
        \hline
    \end{tabular}
    \caption{Utterance-level DistilBERT performance.} % MODIFICATION: Shortened caption
    \label{tab:post_level_module_metrics}
\end{table}

\begin{table*}[ht!]
\normalsize
    \centering
    \Huge
    \begin{adjustbox}{width=1.0\textwidth}
        \begin{tabular}{c|c|c|c|c|c}
             & Method & Precision & Recall & F1 & PRC AUC \\
            \hline \hline\rule{0pt}{12pt}
            \multirow{8}{*}{\rotatebox[origin=c]{90}{Echo}}
            & DeGroot's Diffusion & $0.472 \pm 0.389$ & $0.255 \pm 0.261$ & $0.320 \pm 0.310$ & $0.319 \pm $0.0 \\
            & Node2Vec & $0.764 \pm 0.112$ & $0.819 \pm 0.148$ & $0.788 \pm 0.121$ & $0.806	\pm $0.092 \\
            & GCN & $0.597 \pm 0.192$ & $0.950 \pm 0.041$ & $0.717 \pm 0.127$ & $0.819 \pm 0.119$ \\
            & GraphSAGE & $0.661 \pm 0.097$ & $0.921 \pm 0.029$ & $0.767 \pm 0.076$ & $0.845 \pm 0.113$ \\
            & GAT & $0.755 \pm 0.048$ & $0.914 \pm 0.054$ & $0.825 \pm 0.025^\dag$ & $0.874 \pm 0.074$ \\
            & AGNN & $0.722 \pm 0.108$ & $0.914 \pm 0.070$ & $0.803 \pm 0.084$ & $0.851 \pm 0.117$ \\
            & Fixed-Threshold & $0.654 \pm 0.063$ & $0.627 \pm 0.095$ & $0.633 \pm 0.040$ & $0.678 \pm 0.052$ \\
            & Multimodal Relational Aggregation & $0.789 \pm 0.091$ & $0.878 \pm 0.054$ & $\mathbf{0.826 \pm 0.046}$ & $0.844 \pm 0.076$ \\
            & Distributional (bins) & $0.772 \pm 0.045$ & $0.871 \pm 0.058$ & $0.817 \pm 0.042$ & $0.874 \pm 0.040$ \\
            & Distributional (quantiles) & $0.747 \pm 0.064$ & $0.899 \pm 0.047$ & $0.815 \pm 0.053$ & $0.878 \pm 0.037$ \\
            & Distributional (bins+quantiles) & $0.757 \pm 0.058$ & $0.885 \pm 0.052$ & $0.815 \pm 0.049$ & $0.887 \pm 0.036$ \\
            & Multimodal (Relational + Bins + Quantiles) & $0.766 \pm 0.045$ & $0.892 \pm 0.059$ & $0.824 \pm 0.046^\ddag$ & $0.894 \pm 0.024$ \\
            \hline\hline\rule{0pt}{12pt} 
            \multirow{8}{*}{\rotatebox[origin=c]{90}{Gab}}
            & DeGroot's Diffusion & $0.314 \pm 0.001$ & $0.777 \pm 0.0$ & $0.447 \pm 0.001$ & $0.240 \pm $0.005 \\
            & Node2Vec & $0.404 \pm 0.072$ & $0.335 \pm 0.085$ & $0.364 \pm 0.075$ & $0.386	\pm $0.050\\
            & GCN & $0.296 \pm 0.045$ & $0.678 \pm 0.285$ & $0.387 \pm 0.034$ & $0.344 \pm 0.060$ \\
            & GraphSAGE & $0.390 \pm 0.086$ & $0.469 \pm 0.263$ & $0.387 \pm 0.075$ & $0.397 \pm 0.075$ \\
            & GAT & $0.218 \pm 0.125$ & $0.608 \pm 0.389$ & $0.316 \pm 0.182$ & $0.282 \pm 0.063$ \\
            & AGNN & $0.337 \pm 0.059$ & $0.539 \pm 0.197$ & $0.403 \pm 0.081$ & $0.389 \pm 0.029$ \\
            & Fixed-Threshold & $0.466 \pm 0.072$ & $0.335 \pm 0.061$ & $0.388 \pm 0.061$ & $0.468 \pm 0.093$ \\
            & Multimodal Relational Aggregation & $0.408 \pm 0.067$ & $0.429 \pm 0.094$ & $0.414 \pm 0.069$ & $0.538 \pm 0.070$ \\
            & Distributional (bins) & $0.461 \pm 0.034$ & $0.649 \pm 0.044$ & $0.538 \pm 0.024^\dag$ & $0.521 \pm 0.048$ \\
            & Distributional (quantiles) & $0.429 \pm 0.027$ & $0.702 \pm 0.056$ & $0.532 \pm 0.033$ & $0.533 \pm 0.068$ \\
            & Distributional (bins+quantiles) & $0.435 \pm 0.026$ & $0.714 \pm 0.043$ & $\mathbf{0.540 \pm 0.029}$ & $0.524 \pm 0.064$ \\
            & Multimodal (Relational + Bins + Quantiles) & $0.429 \pm 0.021$ & $0.718 \pm 0.040$ & $0.537 \pm 0.025^\ddag$ & $0.538 \pm 0.070$ \\
            \hline\hline\rule{0pt}{12pt}  
            \multirow{8}{*}{\rotatebox[origin=c]{90}{Parler}}
            & DeGroot's Diffusion & $0.395 \pm 0.221$ & $0.441 \pm 0.247$ & $0.417 \pm 0.233$ & $0.414 \pm $0.001 \\
            & Node2Vec & $0.408 \pm 0.055$ & $0.392 \pm 0.058$ & $0.400 \pm 0.056$ & $0.434 \pm $0.039 \\
            & GCN & $0.284 \pm 0.054$ & $0.760 \pm 0.404$ & $0.348 \pm 0.130$ & $0.410 \pm 0.184$ \\
            & GraphSAGE & $0.309 \pm 0.092$ & $0.649 \pm 0.189$ & $0.394 \pm 0.028$ & $0.343 \pm 0.083$ \\
            & GAT & $0.379 \pm 0.051$ & $0.731 \pm 0.164$ & $0.488 \pm 0.013^\dag$ & $0.315 \pm 0.111$ \\
            & AGNN & $0.369 \pm 0.081$ & $0.552 \pm 0.255$ & $0.416 \pm 0.082$ & $0.449 \pm 0.100$ \\
            & Fixed-Threshold & $0.470 \pm 0.050$ & $0.369 \pm 0.040$ & $0.412 \pm 0.035$ & $0.481 \pm 0.044$ \\
            & Multimodal Relational Aggregation & $0.517 \pm 0.057$ & $0.523 \pm 0.060$ & $\mathbf{0.519 \pm 0.056}$ & $0.382 \pm 0.086$ \\
            & Distributional (bins) & $0.284 \pm 0.037$ & $0.500 \pm 0.075$ & $0.362 \pm 0.049$ & $0.300 \pm 0.031$ \\
            & Distributional (quantiles) & $0.324 \pm 0.018$ & $0.734 \pm 0.057$ & $0.449 \pm 0.025$ & $0.314 \pm 0.030$ \\
            & Distributional (bins+quantiles) & $0.324 \pm 0.021$ & $0.738 \pm 0.051$ & $0.450 \pm 0.027$ & $0.308 \pm 0.031$ \\
            & Multimodal (Relational + Bins + Quantiles) & $0.310 \pm 0.018$ & $0.874 \pm 0.038$ & $0.457 \pm 0.023^\ddag$ & $0.382 \pm 0.086$ \\
        \end{tabular}
    \end{adjustbox}
    \caption{5-Fold CV results on the test sets of Echo, Gab, and Parler datasets using the best (F1-score-wise) configuration. Top results are in bold face, second and third-best results are marked with $\dag$ and $\ddag$, respectively.
    }
    \label{tab:user_level_results}
\end{table*}

\begin{figure*}[ht!]
    \centering
    \includegraphics[width=\textwidth]{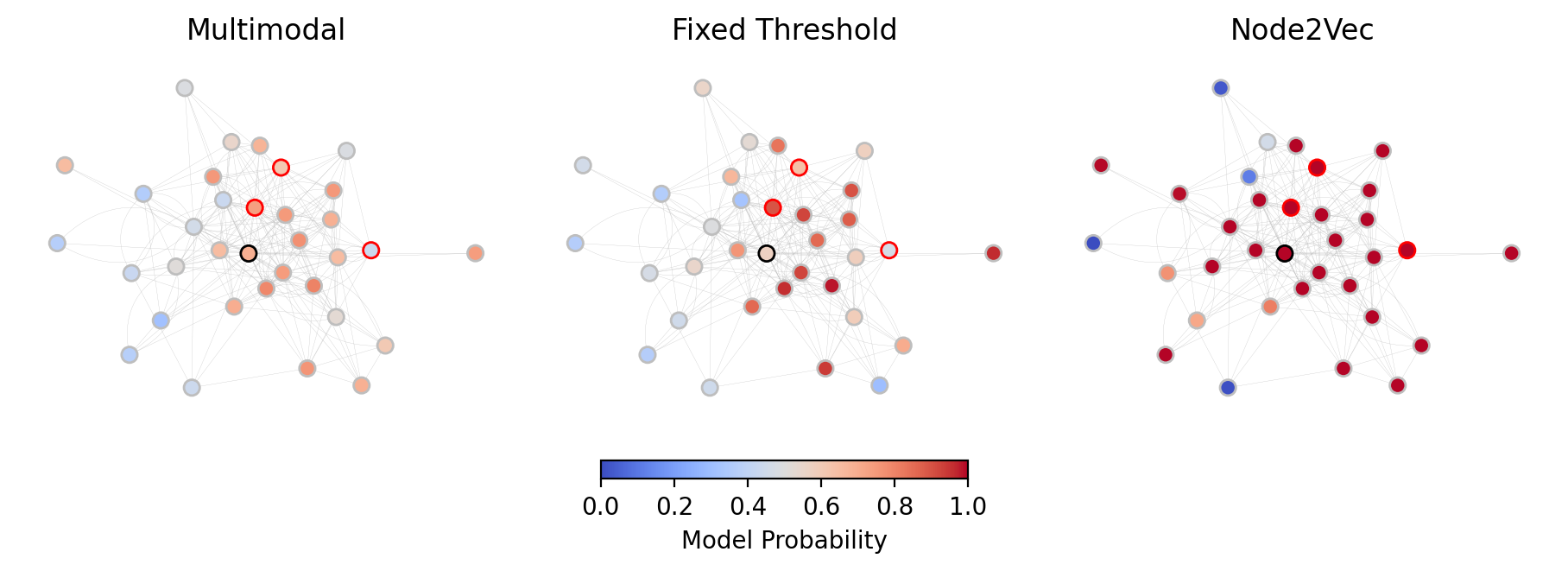} % MODIFICATION: Consider width=0.9\textwidth or similar if needed.
    \caption{Ego network of user 19543095 (Echo data, gold frame=hateful). Node colors show hate probability assigned by: Multimodal (left), Text-only (Fixed-Threshold, middle), Network-only (Node2Vec, right). 0.5 is threshold.} % MODIFICATION: Shortened caption.
    \label{fig:ego_networks_uid1954309514}
\end{figure*}

\paragraph{\bf User level results} Detailed results are presented in Table \ref{tab:user_level_results}. Our Multimodal Relational approach achieves top F1 performance on Echo and Parler. Our combined Distributional (bins+quantiles) approach achieves the best results on Gab, with our Combined Multimodal approach very close third. All proposed aggregative methods (Relational, Distributional variants, Combined) consistently perform competitively, often outperforming the GNN and diffusion models. % MODIFICATION: Condensed results summary.

The optimal weights for the Relational method ($\alpha, \beta, \gamma$ in Eq. \ref{eq:relational}) varied across datasets (more details are available in Appendix \ref{ap:exp_details}), highlighting the influence of platform-specific network structure and dynamics, also reflected in Figure \ref{fig:post_mean_probas_per_user_kde}. % MODIFICATION: Moved specific weights to appendix ref.

\paragraph{Structure-based Analysis}
\label{subsub:strucural}
The F-scores achieved by different aggregation methods show that our approaches consistently outperform baseline algorithms. The multimodal approach performs best on the Echo and Parler datasets, while results differ on the Gab dataset (Table \ref{tab:user_level_results}), suggesting that Gab's size, structure, and unique characteristics influence aggregation effectiveness.

Table \ref{tab:table_datasets4} highlights these differences. The Echo network, the smallest of the three (3.7K users, 20.7K edges), has a clustering coefficient of 0.19 and a degree exponent ($\gamma$) of 2.8, indicating a classic scale-free structure. This topology supports the multimodal approach by effectively leveraging diverse signals. Parler, the largest network (643K users, 11.4M edges), also exhibits a scale-free structure ($\gamma=2.14$) and a higher clustering coefficient (0.224), allowing it to benefit from similar aggregation strategies.
In contrast, Gab (51.2K users, 2.47M edges) has a high clustering coefficient (0.402) and a $\gamma$ of 4.06, indicative of a random-like network structure \cite{amaral2000classes,goh2001universal}. This random-like structure hampers the performance of the multimodal aggregation. However, the Combined Multimodal Aggregation model mitigates this by assigning weights to sub-models, nearly matching the best performerming model.

Breaking down the combined model into its components (Relational, Distributional-bins, Distributional-quantiles) we observe that these methods are competitive, outperform the baselines, and capture the diverse facets of user behavior and network structure. For instance, the Relational method’s performance varies with platform-specific parameters like $\alpha$, $\beta$, and $\gamma$ (see further details in Appendix \ref{ap:exp_details}).

These results underscore the importance of properly addressing the network structure, user dynamics, and platform norms in determining aggregation method effectiveness: distinct structural features, such as clustering coefficients and degree distributions, impact performance, highlighting the need for  (social-) context-specific modeling.

\paragraph{\bf A concrete illustrative example} The benefits of the multimodal approach are evident through the actual probability models assigned to a user.\footnote{The results reported in Table \ref{tab:user_level_results} are based on a binary classification (convert probability to class assignment using the standard 0.5 decision threshold).}
An illustrative example is provided in Figure \ref{fig:ego_networks_uid1954309514}.  
The figure depicts three ego networks of a specific user (black frame; gold label: hate-monger). %Node colors indicate the hate probability assigned to that node by the specific model.
The three ego networks, identical in nodes and structure, differ only in node colors: the color of a node reflects probability assigned by a specific model: Multimodal (left), Text-only (Fixed-Threshold; middle), and network only (node2vec; right). The figure highlights three important observations:
\begin{enumerate}
    \item The text-only aggregative model (Fixed Threshold, optimized) fails to identify the ego node as a hatemonger. The threshold of the text-based aggregative model is optimized in order to achieve the best performance on the data. However, for this reason it can misclassify users that do not cross this threshold. For example, two of the texts posted by the focal node in Figure \ref{fig:ego_networks_uid1954309514} are ``We should take time to thank the left. Every time they call a totally reasonable person a White Supremacist Nazi, you grow our ranks.'' and ``We've done it goys, the \#AltRight has gone mainstream, and we will purge these c*cks from the GOP in the near future''. These (especially the latter one) are clear demonstrations of hate speech. The need for a threshold in text-based models can be justified by the need to give users some level of grace and/or assume that some texts are ironic.
    \item The node2vec classifier indeed classifies the ego-node correctly, but it also significantly overestimates the probabilities of the other nodes, classifying most nodes as hatemongers with very high probability, while classifying only a small subset of nodes as non-hateful users with probabilities close to 0.
    \item Unlike the very high/low probabilities assigned by the network only (node2vec) model,  both the text-only model and the multimodal models assign probabilities closer to the 0.5 threshold. However, while the fine-tuned and calibrated text-only model still fails to cross the decision threshold, the multimodal approach used the network structure to push the decision beyond the threshold, achieving the correct classification.
\end{enumerate}

\paragraph{\bf Further Analysis} While out of the scope of this work, it is worth noting that some models fluctuate in their performance. 

The DeGroot model never achieves top performance. However, while achieving decent results over the GAB and the Parler datasets, it performs poorly over the Echo dataset (F-score: 0.31), compared to the best-performing model (F-score: 0.86) and even to the penultimate GAT (F-score: 0.574). 

Similarly, the node2vec model shows decent results over the Echo and Parler datasets while performing poorly over the Gab dataset with an F-score of 0.364, compared to 0.55 and 0.264 of the best and worst performing models (extended Multimodal and Graph Attention networks, respectively). The performance of the GAT model on the Gab data is also exceptionally poor compared to decent GAT performance over other datasets.  

We attribute these inconsistent performances to the marriage of the unique mechanisms employed by each of the models and the unique characteristics of each dataset in terms of size, density, network structure, betweenness, modularity, etc. (see Tables \ref{tab:table_datasets2} and \ref{tab:table_datasets4}, Figure \ref{fig:post_mean_probas_per_user_kde}, and further details in Appendix \ref{ap:networks}).

These differences may stem from the way the data were collected, as well as different norms and dynamics that emerge on different platforms. 
A careful study of the ways these traits interact and impact model performance is planned for future work. Yet, these fluctuations in the performance of algorithms of different modalities highlight the benefits of our multimodal approach.

\section{Conclusion}
\label{sec:conclusions}
In this work, we proposed a robust and efficient multimodal aggregative method, combining text and social context through relational and distributional aggregations. We demonstrated the benefits of this approach for the task of hate speech and hatemonger detection over three unique and very different datasets from three social platforms: X (Twitter), Gab, and Parler, which we curated ourselves as part of this research and for future studies to come. 
We evaluated our methods on both a large and unique corpus constructed around an ambiguous antisemitic meme and two other datasets based on platforms known for their free speech agenda, and demonstrated their ability to correctly detect hate mongers across different social platforms.

We showed how analyzing both the content and the network features of the users significantly improves the ability to detect hate mongers and provides insights on the promotion of hate. Our findings underscore the importance of a multi modal approach in tackling the complex issue of online hate speech. By integrating textual analysis with social network dynamics, our method achieves higher accuracy and robustness compared to traditional text-only models. 
Furthermore, our approach is scalable and adaptable to various social media platforms, making it a valuable tool for researchers and practitioners aiming to monitor and mitigate hate speech online. We believe that this method can assist in early detection and intervention strategies, ultimately contributing to safer online communities.

\section{Limitations and Ethical Considerations} 
\label{sec:limitations}
\paragraph{Limitations} This work has a number of limitations:
(i) All three aggregation procedures depend on the basic classifier $\theta(t)$. An evasive user who is careful with their words, using only coded language or consistent gas-lighting, may not be identified at all if $\theta(t) < \tau^T$ for all $t \in T^u$ for a user $u$. This reliance on the basic classifier means that if the classifier fails to flag any of the user's texts as problematic, the aggregation methods will also fail to detect the user as a hate monger. This limitation highlights the need for more sophisticated classifiers that can detect subtle or coded forms of hate speech. 
Enhancing the basic classifier's ability to recognize implicit or veiled hate speech, possibly by incorporating contextual or semantic analysis techniques, might deal with that.

(ii) We use the fixed threshold $\tau^U$ as a strong baseline; however, a more comprehensive comparison to other existing methods (e.g., diffusion-based approaches and Graph Neural Networks) should be considered in future work. Although our approach demonstrates effectiveness, benchmarking it against a broader spectrum of state-of-the-art methods would provide a more thorough evaluation of its performance. Additionally, integrating elements from diffusion-based methods or GNNs might enhance our model's ability to capture complex network dynamics and improve detection accuracy.

(iii) Our work focuses on three unique datasets from different social platforms. While unique, these platforms and datasets are mostly associated with users associating themselves with political ideology of the far-right. This ideological tent may limit the generalizability of our findings. Extending the research to include a wider variety of platforms and more diverse datasets could help validate the robustness of our method across different social contexts.

Recognizing the limitations mentioned above is crucial for the ongoing development of more effective methods for detecting hate speech and hate mongers online. Addressing these challenges in future work will be essential for improving detection capabilities and ensuring that methods remain effective against evolving strategies used to spread hate speech.

\paragraph{Ethical Considerations}
The detection of hate mongers on social platforms and the detoxification of discourse are of great importance. However, mislabeling individuals as hate mongers can lead to unintended consequences, including defamation, social ostracization, and psychological distress as well as legal implications. It may also undermine the credibility of the detection systems and erode public trust in moderation efforts. Therefore, while automated systems are valuable for processing large volumes of data efficiently, they should complement rather than replace human judgment.
Furthermore, ethical considerations must guide the deployment of such detection systems. Transparency in how these systems operate and clear communication about the criteria used for labeling are essential. Providing users with avenues to appeal or contest labels can help prevent unjust consequences and maintain trust between the platform and its users.

In this work, we explore computational approaches that can be used to study the phenomena at scale, rather than propose a functional end-to-end content moderation system.

\bibliography{custom_new}

%%%%%%%%%%%%%%%%%%%%%%%%%%%%%%%%%%%%%%%%%%%%%%%%%%%%%%%%%%%%%%%%%%%%%%%%
%%% APPENDICES START HERE %%%%%%%%%%%%%%%%%%%%%%%%%%%%%%%%%%%%%%%%%%%%%%
%%%%%%%%%%%%%%%%%%%%%%%%%%%%%%%%%%%%%%%%%%%%%%%%%%%%%%%%%%%%%%%%%%%%%%%%
\clearpage % MODIFICATION: Ensure appendices start on new page if needed.
\appendix

% MODIFICATION: Added new appendix sections for moved content. Ensure labels are unique.

\section{Analysis of Table 1 Examples}
\label{sec:table1_analysis}
Here we provide a more detailed analysis of the examples presented in Table \ref{tab:hate_texts}:
\begin{itemize}
    \item Utterance \#1: Explicit racism and Islamophobia. Detected by most models, though deHateBERT's confidence (0.57) is surprisingly low.
    \item Utterance \#2: Antisemitism conveyed via the (((echo))) symbol targeting Amy Schumer (who is Jewish). Requires external knowledge of the symbol and target. Missed by DHB and Jigsaw; GPT-4 flags it cautiously.
    \item Utterance \#3: The "14 words," a core white supremacist slogan. Appears innocuous without context. Only detected by the LLMs (GPT-4, Gemini).
    \item Utterances \#4-7: Variations or allusions related to white identity and the "14 words" (\#4, \#5, \#7) or white pride (\#6). Detection is inconsistent. FTDB captures \#4 and \#6 but misses the original \#3. LLMs are uncertain or miss these. \#7 ("White lives matter") co-opts BLM language with white supremacist intent\footnote{\url{www.adl.org/resources/hate-symbol/white-lives-matter}}.
    \item Utterance \#8: "Blue lives matter" responds to BLM, supporting police\footnote{Often used alongside racial slurs but not inherently hate speech itself.}. Not classified as hate speech by the models.
\end{itemize}
These examples highlight the difficulty of detecting coded language, requiring contextual understanding beyond literal text, motivating user-level aggregation.

\section{Prompts and Generated Responses for LLMs}
\label{sec:generated_predictions}
{\bf Offensive content warning:} Some of the prompts (based on the examples in Table \ref{tab:hate_texts}) may be offensive to some readers. Slurs and swear words are obfuscated by using the * symbol. 

% [Content from original Appendix A remains here]
\texttt{Which of the following utterances may be perceived as hate speech?
\begin{enumerate}
    \item Good morning boys and girls!
    \item Muslim terrorists did this and Obama brought them here to run our government. Gotta get rid of Obammy, Omar, and all these other f**king Muslims.
    \item Is there a more debased creature in our modern pop-cultural cesspool than (((Amy Schumer)))? Probably not.
    \item We must secure the existence of our people and a future for white children!.
    \item If white people are going to continue existing - we must support our race.
    \item When the youngest white generation becomes adult they will be 5\% of the world population, down from 1/3 in 1900.
    \item Did you know that most dog breeds were bred by whites? We’re awesome and they’re awesome!
    \item White lives matter!
    \item Blue lives matter!
    \item Black lives matter!
\end{enumerate}
}
\paragraph{\bf GPT-4}
The utterances that may be perceived as hate speech are those that explicitly or implicitly promote hate, discrimination, or hostility towards particular groups based on their ethnicity, religion, or other inherent characteristics. Based on the examples you've provided, here's an analysis of which might be considered hate speech:
\begin{enumerate}
    \item[2.] \textbf{"Muslim terrorists did this and Obama brought them here to run our government. Gotta get rid of Obammy, Omar, and all these other fucking Muslims"} - This statement is clearly hate speech. It uses derogatory language targeting Muslims and makes unfounded, harmful generalizations about an entire religious group.
    \item[3.] \textbf{"Is there a more debased creature in our modern pop-cultural cesspool than (((Amy Schumer)))? Probably not."} - This statement might be perceived as hate speech, particularly because the use of triple parentheses is often a coded way to identify someone as Jewish in an anti-Semitic context.
    \item[4.] \textbf{"We must secure the existence of our people and a future for white children!"} - This phrase is widely recognized as a white supremacist slogan, known as the "Fourteen Words." It is used to promote the idea of white racial superiority and is considered hate speech.
    \item[5.] \textbf{"If white people are going to continue existing - we must support our race"} - This could be interpreted as promoting racial supremacy or separatism, which might be viewed as hate speech depending on the context in which it is used.
\end{enumerate}
The other statements, while potentially controversial or promoting particular views, do not explicitly convey hate against a group based on inherent characteristics in the same way that the identified statements do.
\\
\\
\paragraph{\bf Gemini 1.5 Pro}
\textbf{Analysis of Potentially Hateful Utterances:}
Here's a breakdown of the provided utterances and their potential to be perceived as hate speech:

\begin{enumerate}
    \item \textbf{Good morning boys and girls!}
    \begin{itemize}
        \item This is a simple greeting and does not contain any hateful content.
    \end{itemize}

    \item \textbf{Muslim terrorists did this and Obama brought them here to run our government. Gotta get rid of Obammy, Omar and all these other fucking Muslims.}
    \begin{itemize}
        \item This statement employs harmful stereotypes against Muslims, associating them with terrorism and expressing a desire to remove them from the government. This is a clear example of hate speech.
    \end{itemize}

    \item \textbf{Is there a more debased creature in our modern pop-cultural cesspool than (((Amy Schumer)))? Probably not.}
    \begin{itemize}
        \item The use of triple parentheses around a name is a known anti-semitic symbol online. This statement is likely intended to target Amy Schumer for her Jewish heritage and is therefore hate speech.
    \end{itemize}

    \item \textbf{We must secure the existence of our people and a future for white children!}
    \item \textbf{If white people are going to continue existing - we must support our race}
    \item \textbf{When the youngest white generation becomes adult they will be 5\% of the world population, down from 1/3 in 1900}
    \begin{itemize}
        \item These three statements, while not explicitly hateful, promote a white supremacist ideology. They focus on the preservation of white people and imply a fear of their decline, potentially inciting hostility towards other racial groups.
    \end{itemize}

    \item \textbf{Did you know that most dog breeds were bred by whites? We’re awesome and they’re awesome!}
    \begin{itemize}
        \item This statement, while expressing pride in an accomplishment associated with white people, does not directly attack or demean any other group. It is unlikely to be considered hate speech.
    \end{itemize}

    \item \textbf{White lives matter!}
    \item \textbf{Blue lives matter!}
    \item \textbf{Black lives matter!}
    \begin{itemize}
        \item These statements, on the surface, simply affirm the value of different groups of people. However, depending on the context and intent, they can be interpreted in different ways.
        \item \textit{White lives matter} is often used as a response to the "Black Lives Matter" movement, implying a denial of the systemic racism faced by Black people. In such cases, it can be considered hate speech.
        \item \textit{Blue lives matter} expresses support for law enforcement. While not inherently hateful, it can be used to dismiss or oppose calls for police accountability and racial justice.
        \item \textit{Black lives matter} aims to highlight the systemic racism and violence faced by Black communities. It is a call for justice and equality, not a statement of superiority or hatred towards other groups.
    \end{itemize}
\end{enumerate}

\textit{It's crucial to consider the context and intent behind these statements to determine if they constitute hate speech.}

\section{Additional Experimental Details}
\label{ap:exp_details}
% MODIFICATION: New appendix section for moved content.
\paragraph{Utterance Classification} We used DistilBERT \cite{Sanh2019DistilBERTAD} fine-tuned on each dataset separately. Training used a batch size of 32, AdamW optimizer with learning rate 5e-5, and ran for a maximum of 20 epochs with early stopping based on validation loss (patience 5) using an 80/20 train/validation split. The threshold $\tau^T$ was set to 0.5.

\paragraph{User Classification} All user-level models were evaluated using 5-fold cross-validation on the largest weakly connected component (LCC) of each dataset.
\begin{itemize}
    \item \textbf{GNN Baselines:} We followed the experimental settings from \citet{das2021you} where applicable (e.g., hidden layer sizes, dropout rates). Specific parameters might vary slightly based on library implementations (PyTorch Geometric). Node features for GNNs were derived from aggregating utterance embeddings (e.g., mean pooling of DistilBERT embeddings for user posts).
    \item \textbf{DeGroot's Diffusion:} We followed \citet{israeli2022free}, running for 10 iterations. Initial beliefs were set based on a small seed set (e.g., 5% highest-scoring users based on $\Theta_F$) assigned belief 1 (hateful) or 0 (non-hateful).
    \item \textbf{Node2Vec:} Used implementation from \citet{btab2021}. Parameters: 50 epochs, embedding dimension 128, $p=1, q=1$, 10 walks per node, walk length 20, window size 10. Embeddings were fed into a Logistic Regression classifier.
    \item \textbf{Our Aggregative Methods:}
        \begin{itemize}
            \item Fixed-Threshold ($\Theta_F$): The threshold $\tau^U$ was optimized via grid search on the validation set for each fold (e.g., $\tau^U \in \{1, 3, 5, 10, 20, 50, 100\}$). Table \ref{tab:user_level_results} reports performance with the best $\tau^U$.
            \item Relational ($\Theta_R$): Weights $\alpha, \beta, \gamma$ (Eq. \ref{eq:relational}) were learned using Logistic Regression on features derived from $C_F(u)$ and neighbor hate percentages. Best average weights across folds were approximately: Echo ($\alpha=0.6, \beta=0.8, \gamma=1.5$), Gab ($\alpha=0.8, \beta=0.1, \gamma=0.1$), Parler ($\alpha=0.2, \beta=0.3, \gamma=0.2$).
            \item Distributional ($\Theta_D$): We used $k=10$ bins. For bin-based, bins were [0, 0.1), [0.1, 0.2), ..., [0.9, 1.0]. For quantile-based, the range $[\min \theta(t), \max \theta(t)]$ for user $u$ was divided into 10 equal bins. Features were the normalized counts in each bin (using softmax $\sigma$). Weights $w_i$ (Eq. \ref{eq:distributional}) were learned via Logistic Regression.
            \item Combined Multimodal ($\Theta$): Features from $\Theta_R$, $\Theta^b_D$, and $\Theta^q_D$ were concatenated and fed into a Logistic Regression classifier.
        \end{itemize}
\end{itemize}

\section{Details for Figure 2 Illustrative Example}
\label{ap:fig2_details}
% MODIFICATION: New appendix section for moved content.
The focal user (19543095) in Figure \ref{fig:ego_networks_uid1954309514} was labeled as hateful in the ground truth. The text-only aggregative model ($\Theta_F$ with optimized $\tau^U$) assigned a score below the threshold, misclassifying the user. This occurred despite the user posting clearly hateful content, such as:
\begin{itemize}
    \item ``We should take time to thank the left. Every time they call a totally reasonable person a White Supremacist Nazi, you grow our ranks.'' (Implicitly aligns with those labeled Nazis/Supremacists)
    \item ``We've done it goys, the \#AltRight has gone mainstream, and we will purge these cucks from the GOP in the near future'' (Explicit use of antisemitic slur 'goys', alt-right identification, violent rhetoric 'purge')
\end{itemize}
The need for a threshold $\tau^U > 1$ in text-based models (to account for potential irony, single mistakes, or low-confidence predictions) can lead to missing users like this if their aggregated score doesn't cross the optimized threshold. The network-only model (Node2Vec) correctly classified the user but assigned extremely high probabilities ($>0.9$) to most neighbors, lacking nuance. The multimodal model successfully integrated the weak textual signal (below threshold $\tau^U$) with the strong network signal (hateful neighbors) to correctly classify the user with a more calibrated probability (just above 0.5).

\begin{table*}[h] % MODIFICATION: Relabeled from original Table 6 to avoid conflict.
    \centering
    {\small
        \begin{tabular}{c@{\quad}|c@{\quad}c@{\quad}c@{\quad}|c@{\quad}c@{\quad}}
          {Dataset} & {\#Posts} & {\#Users} & {\#Edges} & {\#Connected Comp.} & {\#Singletons}\\
          \hline\rule{0pt}{12pt}
            Echo (Twitter) & 18M & 7.1K & 21.4K & 9.1K & 2.9K\\ % Rounded
            Gab & 19.4M & 61.4K & 2.63M & 40K & 10.1K\\ % Rounded
            Parler & 115M & 3.1M & 11.1M & 5.5M & 2.4M \\ % Rounded
          \hline
        \end{tabular}
    }
    \caption{Statistics for the full networks (users with >=1 post).}
    \label{tab:table_datasets3_app}
\end{table*}

\section{Networks Statistics}
\label{ap:networks}
% [Content from original Appendix B remains here, including Tables 6 and 7 - relabeled here if needed]
Tables \ref{tab:table_datasets3_app} and \ref{tab:table_datasets4} (in main text) provide network statistics. Table \ref{tab:table_datasets3_app} shows details for the full networks (excluding users with no posts), highlighting the large number of components and singletons, justifying the focus on the largest connected component (LCC) for graph-based analyses (Table \ref{tab:table_datasets4}).

\section{Annotation Guidelines}
\label{annotation_guidelines}
% [Content from original Appendix C remains here]
Each annotator was assigned 50 Parler accounts, ensuring that each account is being annotated by three students. For each account, annotators were given the user name, self-description of the user's account, and 30 posts published by the user. The 30 posts are sampled out of the following: (i) 15 posts with the highest hate speech prediction values;
%\footnote{The prediction is according to the BERT model introduced by \citet{israeli2022free}}; (ii) Its 10 most recent posts;
(ii) Five random posts published by the account. The annotators showed a reasonable agreement level with an average Cohen's Kappa of 0.36. The full guidelines for the Parler dataset annotation process can be found here:
\url{https://shorturl.at/dMwfK}

\section{Full Data and Code} % MODIFICATION: Renamed from Coda
\label{ap:data_code}
% [Content from original Appendix D remains here]
The full data and code can be found here: \url{http://bit.ly/4gMG3bl}

\end{document}